\pdfoutput=1

\documentclass[11pt]{article}

\usepackage{coling}

\usepackage{times}
\usepackage{latexsym}
\usepackage{graphicx}
\usepackage{xcolor}
\usepackage[T1]{fontenc}

\usepackage[utf8]{inputenc}

\usepackage{microtype}
\usepackage{booktabs} 
\usepackage{multirow} 
\usepackage{siunitx} 
\usepackage{amsmath}
\usepackage{latexsym}
\usepackage{amssymb}
\usepackage{amsmath}
\usepackage{amsthm}
\usepackage{booktabs}
\usepackage{enumitem}
\usepackage{natbib}
\usepackage{color}
\usepackage{times}
\usepackage{latexsym}

\usepackage[T1]{fontenc}
\usepackage[utf8]{inputenc} 
\usepackage{multirow} 
\usepackage{siunitx} 
\usepackage{microtype}
\usepackage{inconsolata}
\usepackage{tabularx}
\usepackage{colortbl} 
\usepackage{siunitx} 
\usepackage{inconsolata}

\title{General LLMs as Instructors for Domain-Specific LLMs: A Sequential Fusion Method to Integrate Extraction and Editing}

\author{Xin Zhang\textsuperscript{1}, Tianjie Ju\textsuperscript{2}, Huijia Liang\textsuperscript{3}, Ying Fu\textsuperscript{4}, Qin Zhang\textsuperscript{1}\thanks{~~Corresponding author}$^{\dagger}$ \\
\textsuperscript{1}Central South University \\
 \textsuperscript{2}Shanghai Jiao Tong University \\
 \textsuperscript{3}SiChuan University \\
 \textsuperscript{4}Xidian University \\
  \texttt{\{ZhangXin221@csu.edu.cn,jometeorie@sjtu.edu.cn,huijia.leung@foxmail.com\}} \\
  \texttt{\{fuying@stu.xidian.edu.cn,qqzhang11@126.com\}}}

\begin{document}
\maketitle
\begin{abstract}
The substantial interest in updating Large Language Models (LLMs) without retraining from scratch is accompanied by several challenges. This is particularly true when updating LLMs with datasets that necessitate domain-expert reasoning across extensive texts, despite limited samples. We termed the scenario as the Few-Shot Domain-Expert Reasoning for Updating LLMs (FDoR-UL). Traditional methods such as Low-Rank Adaptation (LoRA) and Retrieval Augmented Generation (RAG) are inadequate for addressing this critical issue, particularly evident in our exploration of a specific medical dataset that epitomizes the distinct needs of FDoR-UL. To tackle this challenge, we introduce a Sequential Fusion method to integrate knowledge from complex contexts into LLMs. This method employs a two-stage framework: initially leveraging general LLMs to perform relation extraction for knowledge acquisition from complex texts, followed by updating domain-specific LLMs through Knowledge Editing (KE). Employing our method, domain-specific LLMs achieved a 71.7\% accuracy (an average gain of 39.1\%) in question-answering tasks. Furthermore, we expanded our evaluation to a novel economics-management dataset we developed, where our method achieved a 75.0\% accuracy (an average gain of 45.0\%). These findings underscore the effectiveness and flexibility of our approach in FDoR-UL across various domains. The code and dataset will be released in the revised version.
\end{abstract}

\section{Introduction}
Large language models (LLMs) such as GPTs~\citep{DBLP:journals/corr/abs-2303-08774} and LLaMA~\citep{DBLP:journals/corr/abs-2307-09288} have showcased their remarkable ability in language understanding and knowledge assimilation, attributable to extensive training on vast text datasets~\citep{DBLP:conf/emnlp/MishraTC22, DBLP:journals/corr/abs-2303-18223}. Despite their prowess, LLMs face considerable challenges in staying abreast with the evolving landscape of world knowledge and specialized domain insights. The limitations stem from reliance on potentially outdated or inaccurate datasets, leading to issues like erroneous outputs and hallucinations~\citep{DBLP:conf/emnlp/DongDSXSL22, DBLP:conf/acl/RashkinRT020, DBLP:journals/tacl/DhingraCEGEC22}. Concurrently, there is an escalating demand for these models to incorporate the most recent information or private, industry-specific data~\citep{DBLP:journals/corr/abs-2304-06975,DBLP:journals/corr/abs-2309-11325}.

\begin{figure}[t]
\centering 
\includegraphics[width=0.5\textwidth]{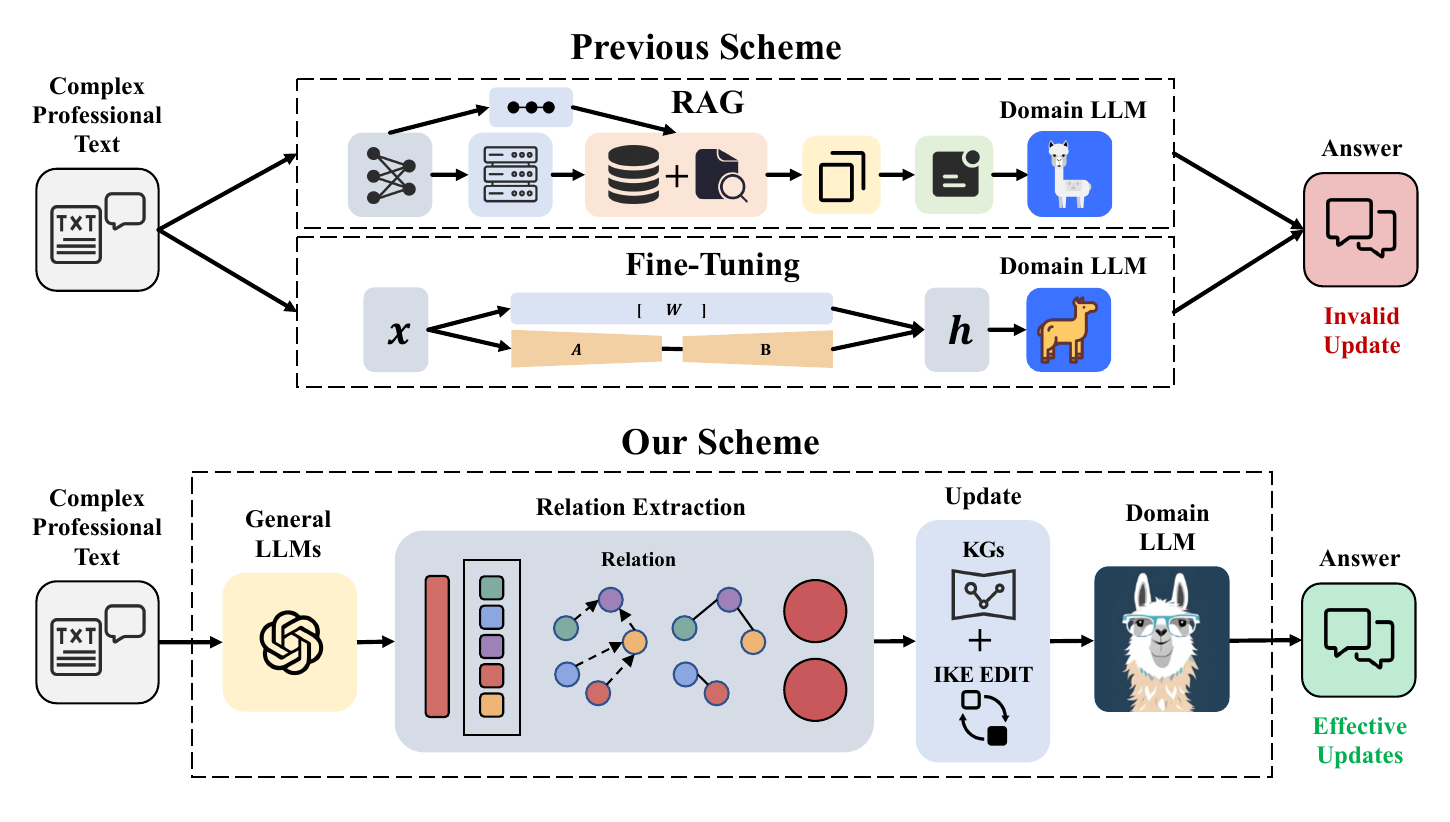} 
\caption{An illustration of our proposed Sequential Fusion method, showcasing its comparison with traditional approaches. Our method employs a two-stage framework: First, general LLMs are leveraged to perform relation extraction for extracting structured knowledge from complex texts. Second, the extracted knowledge is used to update domain-specific LLMs through knowledge editing.} 
\label{sec:contrast}
\end{figure} 
\par
Although previous studies have made strides in updating LLMs through methods like fine-tuning~\citep{DBLP:conf/iclr/HuSWALWWC22,DBLP:conf/acl/LiL20} and Retrieval Augmented Generation (RAG)~\citep{DBLP:conf/nips/LewisPPPKGKLYR020,DBLP:conf/coling/MathurLLMKAMZ24}, our investigations into Drug Combo Extraction (DCE) within the FDoR-UL framework have encountered typical challenges associated with these approaches.
Specifically, Low-Rank Adaptation (LoRA) struggles with effectuating meaningful updates from a limited samples~\cite{DBLP:journals/corr/abs-2306-07967}. Similarly, RAG is prone to experience hallucinations when dealing with intricate semantic materials~\citep{DBLP:conf/nips/LewisPPPKGKLYR020}.\par
Furthermore, certain domain-specific LLMs have demonstrated deficiencies in performing complex reasoning tasks, which can be attributed to limitations in their model parameters and scale. 
These observations necessitate the development of tailored solutions that directly confront and mitigate these shortcomings in FDoR-UL. To mitigate the shortcomings in FDoR-UL experienced by LoRA and RAG, we propose the Sequential Fusion. Its benefits are detailed in Figure~\ref{sec:contrast}, showcasing the method's enhancements over traditional mechanisms.
\par
Initially, we employ LLMs for relation extraction from complex texts, facilitating effective knowledge acquisition. Subsequently, LLMs are refined by integrating extracted knowledge via the Knowledge Editing (KE) method IKE~\cite{DBLP:conf/emnlp/ZhengLDFWXC23}. This method was selected as the preferred approach following comprehensive comparative experiments that assessed different update methods and knowledge representations in complex texts. Through IKE, knowledge extracted in the first stage can be applied to update the domain-specific LLMs.
\par
Our method has shown promising results in FDoR-UL, achieving a 71.7\% accuracy in question-answering tasks within the domain of DCE. To further validate our method and provide a more robust evaluation, we proposed the Moderating Effect-Extraction (MEE). This dataset serves as a benchmark in the fields of economics and management, focusing on the nuances of interaction effects. By applying our method to the MEE, we attained a 75.0\% accuracy in question-answering tasks. Specifically, the average growth rate reached 39.1\% on DCE and an impressive 45.0\% on MEE. 
\par
We have made notable improvements in the challenge of FDoR-UL. Compared to previous methods LoRA and RAG, our strategy represents a significant advancement in this field. In summary, the contributions of this work are fourfold:

\begin{itemize}
    \item We are the first to identify the FDoR-UL scenario and propose an effective two-stage framework tailored for efficient adaptation in this setting..
    \item We introduced MEE as a new benchmark for FDoR-UL, also offering a new choice for LLMs performing sequence-to-sequence relation extraction.
    \item Initially focusing on KE in complex texts, our comparative experiments uncovered new features across different knowledge representations and update methods. This enabled us to propose a reference update strategy for the second stage.
    \item Our approach exhibits a substantial enhancement over LoRA and  RAG on DCE and our internally developed MEE, positioning it as the currently optimal solution in the field.
\end{itemize}

\section{Related Work}

\subsection{LLMs performing relation extraction}
In recent years, LLMs represented by GPT have driven a paradigm shift in relation extraction, transitioning from traditional discriminative approaches to generative methods ~\citep{DBLP:journals/corr/abs-2312-17617}. This shift reframes relation extraction as a generative task, allowing models to output relational information in triplet format directly, which reduces dependency on fixed label sets and enhances flexibility and generalization.

~\citet{DBLP:conf/emnlp/WanCMLSLK23} introduced a contextual learning approach in this field. Unlike traditional supervised methods that require extensive annotated data, GPT-RE performs relation extraction using task prompts with a few examples, achieving inference with GPT-3. \citet{DBLP:conf/acl/WadhwaAW23} advanced this approach by linearizing relation triplets, enabling generative models to output relations as generated strings. Initial experiments showed that Flan-T5 underperformed compared to GPT-3 in few-shot learning, though Chain of Thought (CoT) fine-tuning improved its performance significantly. ~\citet{DBLP:conf/acl/ZhangG023} further redefined relation extraction as a question-answering task with the QA4RE framework, using instruction tuning to leverage generative models’ strengths and enhance zero-shot performance in GPT-3.5 and Flan-T5.

For document-level relation extraction, ~\citet{DBLP:conf/emnlp/LiJZ23} proposed a method combining large models with a Natural Language Inference (NLI) module to handle complex relations in longer texts, utilizing LLMs’ generative capacity alongside NLI’s logical reasoning. ~\citet{DBLP:conf/www/SunHYTZP24} introduced GenRDK, a framework that uses models like ChatGPT and LLaMA to generate synthetic data, guiding the generation of complex, long texts through chain-retrieval prompting to extract relation triplets effectively.

\subsection{Updating LLMs}
To efficiently update LLMs without the need for retraining, methods such as fine-tuning, exemplified by LoRA~\cite{DBLP:conf/iclr/HuSWALWWC22} offer targeted adjustments of model parameters, circumventing the exhaustive retraining process. RAG further enriches LLMs by integrating external knowledge bases, utilizing indexing, retrieval, and generation sequences~\citep{DBLP:conf/nips/LewisPPPKGKLYR020}.
Knowledge Editing (KE), as outlined in recent studies ~\citep{DBLP:conf/iclr/SinitsinPPPB20,DBLP:conf/emnlp/CaoAT21}, offers a novel approach for updating LLMs. KE allows for precise updates to the model's knowledge without affecting its overall integrity. This method encompasses strategies that introduce new components or parameters—such as SERAC ~\citep{DBLP:conf/icml/MitchellLBMF22}, CaliNET ~\citep{DBLP:conf/emnlp/DongDSXSL22}, and GRACE ~\citep{DBLP:conf/nips/HartvigsenSPKG23}—as well as techniques that directly modify existing parameters, demonstrated by meta-learning approaches like MEND ~\citep{DBLP:conf/iclr/MitchellLBFM22}, and targeted modifications through ROME ~\citep{DBLP:conf/nips/MengBAB22} and MEMIT ~\citep{DBLP:conf/iclr/MengSABB23}.

\par
In our exploration of updating LLMs, we present a novel approach that surpasses traditional methods. Unlike standalone techniques such as RAG or fine-tuning, which, while beneficial, struggle in complex, data-scarce settings.Our Sequential Fusion method combines the strengths of relation extraction and targeted knowledge editing, marking a significant advancement in the FDoR-UL scenario.

\section{Method}
Reflecting the two-stage approach detailed previously, our method section is structured around initially extracting knowledge through RE and subsequently applying the knowledge to improve domain-specific LLMs.
\subsection{Relation Extraction}
\label{sec:RE}
\subsubsection{Task Formulation}
In the framework of LLMs, the task of RE is treated as a conditional sequence-to-sequence problem, described through the following mathematical formulation. Given a complex text \(C\) and a prompt \(x\), the objective is to generate a sequence \(y\) that accurately represents the extracted relations. The probability of generating the entire sequence \(y\), conditioned on the complex text \(C\) and the prompt \(x\), is denoted by \(p_{\text{LM}}(y | C, x)\). This probability can be decomposed into the product of conditional probabilities for each token \(y_t\) in the output sequence \(y\), \(T\) represents the number of tokens in the output sequence as shown in the equation below:
\begin{equation}
p_{\text{LM}}(y|C,x) = \prod_{t=1}^{T} p(y_t|C,x,y_{1:t-1})
\end{equation}
\subsubsection{Prompt Strategy}
\begin{figure*}
\centering
\includegraphics[width=\textwidth]{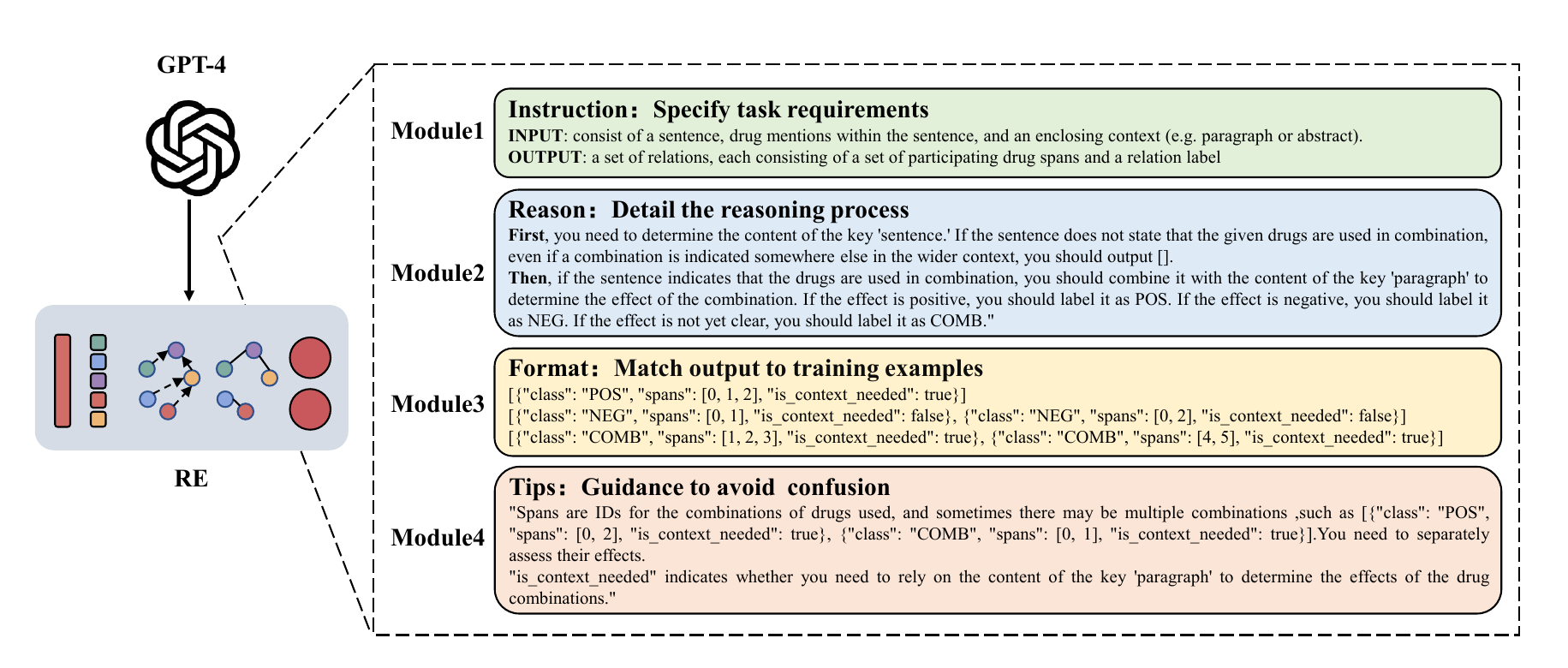}
\caption{An overview of our prompt strategy, organized into the INSTRUCTION, REASON, FORMAT, and TIPS modules, for the relation extraction task.}
\label{sec:prompt}
\end{figure*}

The RE task between DCE and MEE demands the model to deeply comprehend text, apply reasoning, and discern relevant entities and their relationships. In response to this imperative, our prompting strategy, as illustrated in Figure~\ref{sec:prompt}, comprises several key components designed to guide the model through this intricate process.\par
The strategy begins with the \textbf{INSTRUCTION Module}, which clarifies the specific requirements of the relation extraction task, ensuring the model fully understands the directives. Following this, the \textbf{REASON Module} integrates intermediate reasoning steps into the prompt, facilitating a deeper understanding of the task at hand. Concurrently, the \textbf{FORMAT Module} provides sample outputs to help align the model's responses with the structure of the training examples, thus enhancing the consistency and relevance of the output. Additionally, the \textbf{TIPS Module} offers clarifying information in the context of complex texts, assisting the model in navigating potential ambiguities and thereby reducing confusion and improving accuracy. The comprehensive prompts used on the DCE and MEE are detailed in the Appendix~\ref{prompt}.

\subsection{Refocusing LLMs with Extracted Knowledge}
\begin{figure}
\centering
\includegraphics[width=0.5\textwidth]{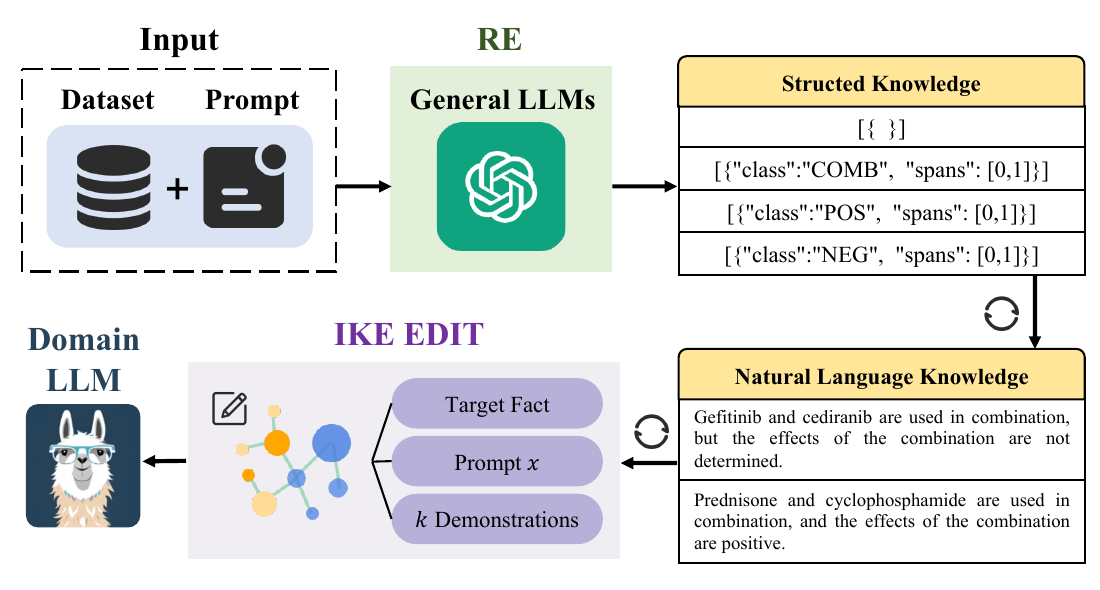}
\caption{ A workflow diagram illustrating the process of updating domain-specific LLMs through Sequential Fusion. This includes extracting structured knowledge, converting it into natural language, and editing the LLMs using the IKE method.}
\label{sec:workflow}
\end{figure}

As shown in Figure~\ref{sec:workflow}, we implemented Sequential Fusion to update domain-specific LLMs. By applying the method outlined in Section~\ref{sec:RE}, we efficiently generated structured knowledge. This knowledge was subsequently converted into natural language format and merged with various natural language demonstrations to form the input data. Following this, we utilized IKE~\citep{DBLP:conf/emnlp/ZhengLDFWXC23} for editing domain-specific LLMs with the input data, as it demonstrated superior editing performance without necessitating parameter adjustments.
Simultaneously, We compared three distinct KE paradigms: IKE(In-Context Editing), ROME(Locate then Edit), and GRACE(Memory-based), alongside traditional methods like FT, LoRA, and RAG. Details follow in the next section.

\par
An example of updating the domain-specific LLM with extracted knowledge is as follows:
Initially, we utilize GPT-4 for relation extraction on DCE, resulting in structured knowledge representations, exemplified by \{\texttt{"class"}: \texttt{"POS"}, \texttt{"spans"}: \texttt{[0,1]}\}. This structured knowledge enhances clarity and facilitates retrieval. However, its contextual information is limited, making it suboptimal for the IKE method.
\par
\begin{figure}
\centering
\includegraphics[width=0.5\textwidth]{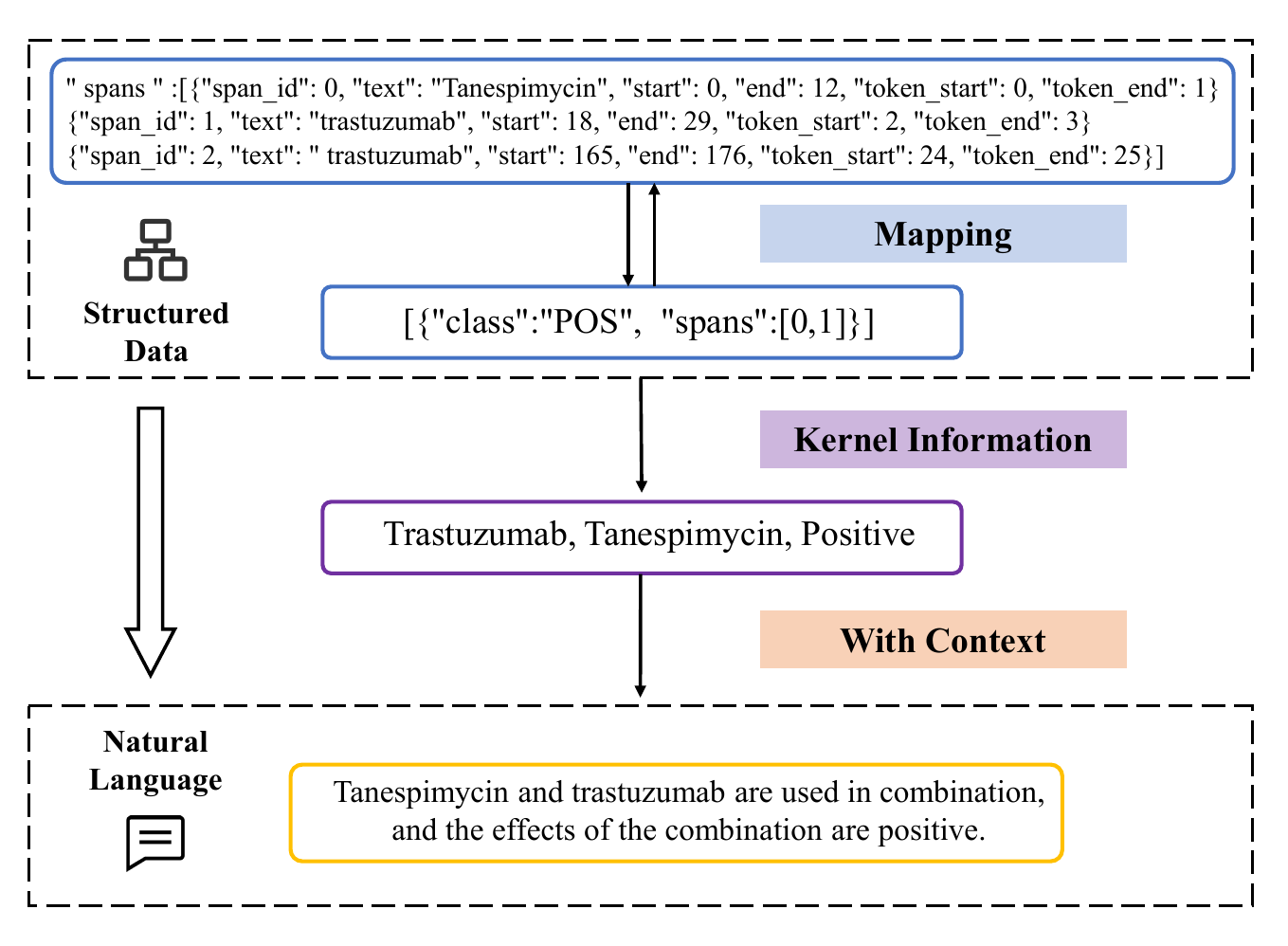}
\caption{The Process of the SKT Module. This diagram illustrates how structured knowledge is converted into natural language through the Mapping and Semantic Integration stages.}
\label{sec:SKT}
\end{figure}

Subsequently, our Structured Knowledge Transformation (SKT) module translates structured knowledge into natural language, rendering relation extraction findings into comprehensible text. Comprising two main components, the specifics of the SKT module are detailed in Figure~\ref{sec:SKT}.
\par
\textbf{Mapping}:
Given the structured knowledge \(K\) with identifiers \(k_i\), and a knowledge base \(D\) containing entities with detailed attributes, we define the mapping function \(M(K, D)\) as:

\begin{multline}
M(K, D) = \{ (d_i^{\text{attr}_1}, d_i^{\text{attr}_2}, \ldots) \mid (k_i \in K) \land \\
(d_i \in D) \land (d_i.\text{id} = k_i) \}
\end{multline}

where \(k_i\) is an identifier in \(K\), and \(d_i\) is the corresponding entity in \(D\) with \(d_i.\text{id} = k_i\). The output of \(M(K, D)\) is a set of tuples, each consisting of selected attributes (\(attr_1, attr_2, \ldots\)) from the entity \(d_i\) in \(D\) that matches the identifier \(k_i\) from \(K\). 
\par
\textbf{Semantic Integration}:
The transformation of structured knowledge into natural language descriptions is formalized as:

\begin{equation}
N = \Phi \left( \bigoplus_{i=1}^n d_i, \Gamma \right)
\end{equation}

where \(\Gamma\) represents the contextual information. When combined with the entities \(d_i\) derived from \(D\) through the mapping process \(M(K, D)\), it enables the creation of natural language description \(N\).
\par
For example, structured knowledge \{\texttt{"class"}: \texttt{"POS"}, \texttt{"spans"}: \texttt{[0,1]}\} is transformed into the narrative statement: "Tanespimycin and trastuzumab are used in combination, and the effects of the combination are positive".

\par
Using such natural language descriptions, we generate question-answer pairs and edit examples to update domain-specific LLMs. Our objective, refined using the IKE edit, is articulated as:
\begin{equation}
\begin{split}
    \max &\left( l_{\text{align}} \cdot p(y^*|x,N,C;\theta) \right) - \\
    \max &\left( (1-l_{\text{align}}) \cdot \mathcal{L}(y,p(y|x,N,C;\theta)) \right)
\end{split}
\end{equation}
In this formulation:
\begin{itemize}
    \item Define the alignment flag \( l_{\text{align}} \) as a metric to ascertain whether a given prompt \( x \) is consistent with the elements of set \( X_{\text{align}} \)
    \item The model's capabilities and performance are defined by demonstrations $C$ and parameters $\theta$, evaluated by $\mathcal{L}(y, p(y|x,N,C;\theta))$.
\end{itemize}

\section{Experimental Setup}
Our experiment follows the outlined method, starting with RE and then testing LoRA, RAG, and Sequential Fusion on two datasets. The supplementary assessment compares IKE, ROME, GRACE, FT, LoRA, and RAG with the extracted knowledge to determine the optimal update approach for the second stage of Sequential Fusion.
\subsection{Relation Extraction}
\subsubsection{Dataset}
\paragraph{DCE} DCE is designed for the systematic extraction of information on known drug combinations and their impacts from scientific texts. This involves processing sentences with drug mentions and contextual information from paragraphs or abstracts to create sets of relations that link drug mentions to a label indicating the effect of their combination. The dataset includes 1,634 annotated abstracts, with 1,362 for training and 272 for testing. The task's complexity and challenges stem from handling variable-length n-ary relations and intricate text-level dependencies. 
\paragraph{MEE} is utilized to explore the moderating effect between economic and management variables through relation extraction from prominent journals in the field. The dataset consists of 100 annotated paragraphs, split into an 80\% training set and a 20\% testing set. To annotate moderating effects in these articles, a categorization scheme similar to that used in DCE was adopted. Effects are classified into four categories: \textbf{POS} (moderating effect strengthens the main effect), \textbf{NEG} (moderating effect weakens the main effect), \textbf{COMB} (interaction or undetermined), and \textbf{NO\_COMB} (no moderating effect). We streamlined the annotation process for LLM-friendly sequence-to-sequence extraction, omitting NER annotations.

Appendix~\ref{prompt} provides examples of DCE and MEE, highlighting their input and output structures as well as their foundational content. Generating accurate outputs requires complex reasoning from extensive texts and integration of domain knowledge.
 
\subsubsection{Evaluation Metrics}
We adhered to the evaluation criteria outlined by \cite{DBLP:conf/naacl/Tiktinsky0NASTH22}, focusing on two scenarios: Partial Match and Exact Match. The metric calculations in the current study employed the label-to-index binary encoding strategy detailed in Appendix~\ref{sec:binary encoding}, as established by the initial research.

\subsubsection{Baselines}
In our analysis, we used PubmedBERT~\cite{DBLP:journals/health/GuTCLULNGP22}, SciBERT~\cite{DBLP:conf/acl/GururanganMSLBD20}, and BioBERT~\cite{DBLP:journals/bioinformatics/LeeYKKKSK20} as baseline models. These models underwent Domain-Adaptive Pre-Training (DAPT)~\cite{DBLP:conf/naacl/Tiktinsky0NASTH22}

\subsection{Experiment on Methods for Updating LLMs}
\subsubsection{Dataset\&Technical Details}
We harnessed inputs from DCE and MEE to serve as a foundational knowledge base for LoRA and RAG. Knowledge extracted from GPT-4 was utilized to generate new datasets, named \textbf{E-DCE} and \textbf{E-MEE}, which were employed as data sources for Sequential Fusion.
Our experiments were conducted using three NVIDIA RTX 4090 GPUs, with Llama2 7b~\cite{DBLP:journals/corr/abs-2307-09288} and Qwen 7b~\cite{DBLP:journals/corr/abs-2309-16609} as the backbone for our evaluation.

\subsubsection{Evaluation Metrics}
In evaluating question-answering with LLMs, accuracy is widely used as a metric. ~\citep{DBLP:conf/iclr/LiZCDJPB24}. However, due to the challenges of DCE and MEE, We refined the metric in Appendix~\ref{Evaluation Metrics}, building on the framework of ~\citep{DBLP:conf/naacl/Tiktinsky0NASTH22} to capture the subtleties of LLMs and dataset complexities. The refinement was supported by preliminary experiments that examined the probabilistic nature of LLM responses.

\subsection{Refocusing LLMs with Extracted Knowledge}
We conducted a comparative experiment of six different methods: IKE, ROME, GRACE, FT, LoRA, and RAG, using the \textbf{E-DCE} and \textbf{E-MEE} to identify the optimal update method for the second phase. Each method was evaluated using two knowledge representation formats: structured and natural language descriptions. The evaluation also employed the refined accuracy metric, complemented by a strategy of manual verification.

\section{Result and Analysis}
Here, we analyze the results from our three experiments, linking each to its distinct implications.
\subsection{Relation Extraction}
Comparative analysis revealed that using the CoT approach, GPT-4 delivered its best performance, as evidenced by the results in Table~\ref{sec:re_result}. During our exploration of RE, we delved into the method of Few-shot prompt and CoT prompt. The findings suggest that improving RE performance in LLMs does not hinge on increasing the number of prompts. Instead, the essence of advancement lies in refining the model's proficiency in understanding and executing the reasoning process.

\begin{table*}[ht]
\caption{A comparison of RE performance utilizing GPT-4, featuring two different prompting strategies versus baselines on the DCE. For the MEE, the comparison is limited to the two GPT-4-based prompting strategies.}
\label{sec:re_result}
\centering
\setlength\tabcolsep{10pt} 
\renewcommand{\arraystretch}{1.3} 
\begin{tabular}{@{}c c c c c c@{}}
\toprule
\multirow{2}{*}{\centering\textbf{Dataset}}& \multirow{2}{*}{\centering\textbf{Method}} & \multicolumn{2}{c}{\textbf{Positive Combination F1}} & \multicolumn{2}{c}{\textbf{Any Combination F1}} \\
\cmidrule(lr){3-4} \cmidrule(lr){5-6}
& & Exact Match & Partial Match & Exact Match & Partial Match \\
\midrule
\multirow{5}{*}{\centering DCE} & PubMed BERT & 61.8 & 67.7 & 69.4 & 77.5 \\
    & BioBERT & 56.0 & 63.5 & 65.6 & 75.7 \\
    & BlueBERT & 56.6 & 63.5 & 64.2 & 74.7 \\
    & CoT Prompt & \textbf{71.4} & \textbf{75.5} & \textbf{78.6} & \textbf{81.3} \\
    & Few-shot Prompt & 50.5 & 55.8 & 70.3 & 74.5 \\
\midrule
\multirow{2}{*}{\centering MEE} & CoT Prompt & \textbf{85.7} & \textbf{85.7} & \textbf{91.9} & \textbf{91.9} \\
    & Few-shot Prompt & 78.8 & 78.8 & 88.9 & 88.9 \\
\bottomrule
\end{tabular}
\end{table*}

\subsection{Experiment on Methods for Updating LLM}
To address the challenge of LLMs generating outputs that are semantically correct but stylistically inconsistent \citep{DBLP:conf/acl/WadhwaAW23}, we engaged domain-expert master's students for manual cross-verification.

\subsubsection{Main Results}
\begin{figure}
\centering
\includegraphics[width=0.5\textwidth]{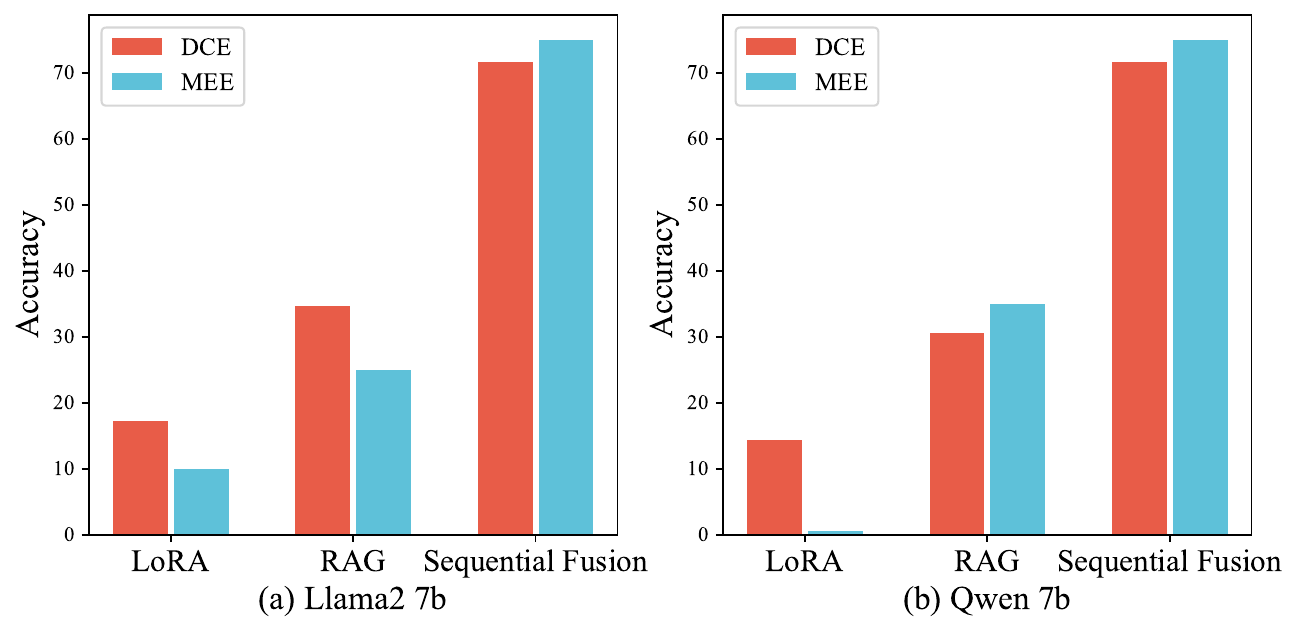}
\caption{A comparison chart illustrating the question-answering accuracy of LoRA, RAG, and Sequential Fusion on the DCE and MEE of Llama2 7b and Qwen 7b.}
\label{main_result}
\end{figure}

Figure~\ref{main_result} shows LoRA and RAG exhibit lower initial accuracy in the question-answering tasks. However, with the Sequential Fusion method, the accuracy of the Llama 2 7b and Qwen 7b models on the DCE and MEE datasets significantly increased to 71.9\% and 75.0\% respectively, clearly outperforming the individual LoRA and RAG methods. Specifically, the models achieved an average growth rate of 39.1\% on DCE and 45.0\% on MEE.
\par
Having established the performance metrics of LoRA, RAG, and Sequential Fusion, we now turn our attention to the types of errors each method exhibits, highlighting their impacts in complex question scenarios. During our manual evaluation,  we classified the errors from various methods in the question-answering task into four types: inability to provide answers, contradictory responses to the same combination, incorrect effects from combinations, and omissions in combined information. 
\par
As detailed in Appendix~\ref{Error Types}, the use of LoRA has not fully overcome the model's reliance on pre-existing knowledge, limiting its ability to deliver definitive answers. Moreover, while the RAG  attempts to enhance responses by incorporating external information, it often results in reasoning errors and hallucinations. In contrast, Sequential Fusion reduces hallucinations and enhances the model's ability to handle complex challenges.

\subsubsection{Case Study}
\begin{figure}
\centering
\includegraphics[width=0.5\textwidth]{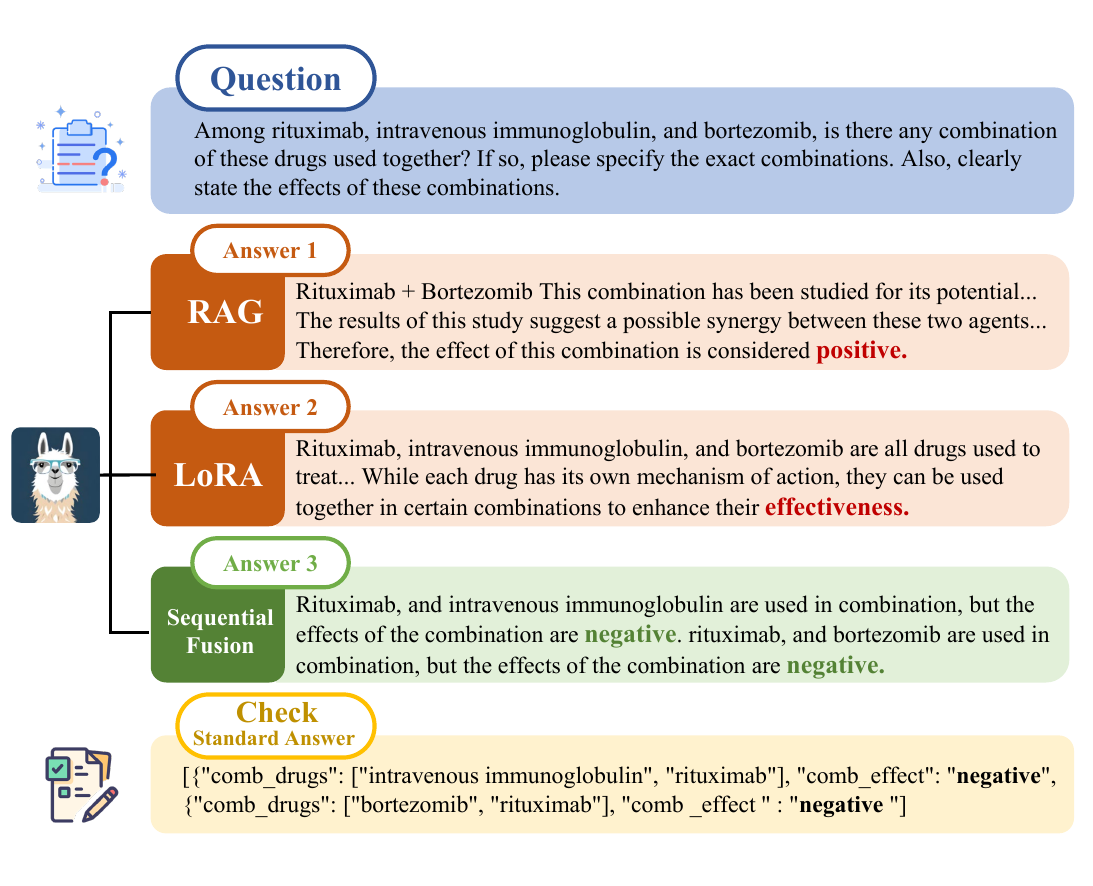}
\caption{A case study visualizing the question-and-answer outcomes of a specific example in the DCE using three methods: LoRA, RAG, and Sequential Fusion.}
\label{case_study}
\end{figure}

To demonstrate the efficacy of our method, we present an example from the update of the Llama2 7b on DCE. As depicted in Figure~\ref{case_study}, the standard model identified two drug combinations, both of which were determined to have negative effects. The RAG identified only one combination and, similar to the LoRA, failed to accurately assess the effects of the drug combinations. In contrast, Sequential Fusion not only precisely identified all drug combinations but also was the sole model to correctly evaluate their effects.

\begin{table*}[ht]
\caption{A Comparison of Model Performance Using Structured Knowledge and Natural Language Representations on E-DCE and E-MEE for Llama2 7b and Qwen 7b. E-DSK and E-DNK refer to structured and natural language representations in E-DCE, while E-MSK and E-MNK denote the same for E-MEE.}
\label{sec:comparative}
\centering
\setlength\tabcolsep{8pt} 
\renewcommand{\arraystretch}{1.1} 
\begin{tabular}{@{}lS[table-format=3.1]S[table-format=2.1]S[table-format=3.1]S[table-format=2.1]S[table-format=3.1]S[table-format=2.1]S[table-format=3.1]S[table-format=2.1]@{}}
\toprule
\multirow{2}{*}{\textbf{Model}} & \multicolumn{4}{c}{\textbf{Llama2 7b}} & \multicolumn{4}{c}{\textbf{Qwen 7b}} \\
\cmidrule(lr){2-5} \cmidrule(lr){6-9} 
 & {\textbf{E-DNK}} & {\textbf{E-DST}} & {\textbf{E-MNK}} & {\textbf{E-MSK}} & {\textbf{E-DNK}} & {\textbf{E-DSK}} & {\textbf{E-MNK}} & {\textbf{E-MSK}} \\
\midrule
FT & 9.9 & 5.0 & 0.0 & 0.0 & 10.3 & 13.2 & 0.0 & 0.0 \\
LoRA & 15.3 & 15.3 & 10.0 & 10.0 & 9.6 & 9.6 & 0.0 & 0.0 \\
RAG & 45.9 & 29.8 & 40.0 & 20.0 & 45.4 & 29.4 & 45.0 & 25.0 \\
GRACE & 20.5 & 20.5 & 15.0 & 15.0 & / & / & / & / \\
ROME & 62.3 & \textbf{71.5} & 80.0 & \textbf{95.0} & 72.7 & \textbf{79.4} & 80.0 & \textbf{90.0} \\

IKE & \cellcolor{blue!25}\textbf{100.0} & / & \cellcolor{blue!25}\textbf{100.0} & / & \cellcolor{blue!25}\textbf{100.0} & / & \cellcolor{blue!25}\textbf{100.0} & / \\
\bottomrule
\end{tabular}
\end{table*}

\subsection{Refocusing LLMs with Extracted Knowledge}
\subsubsection{Comparison between Updating Methods}

\begin{figure}
\centering
\includegraphics[width=0.5\textwidth]{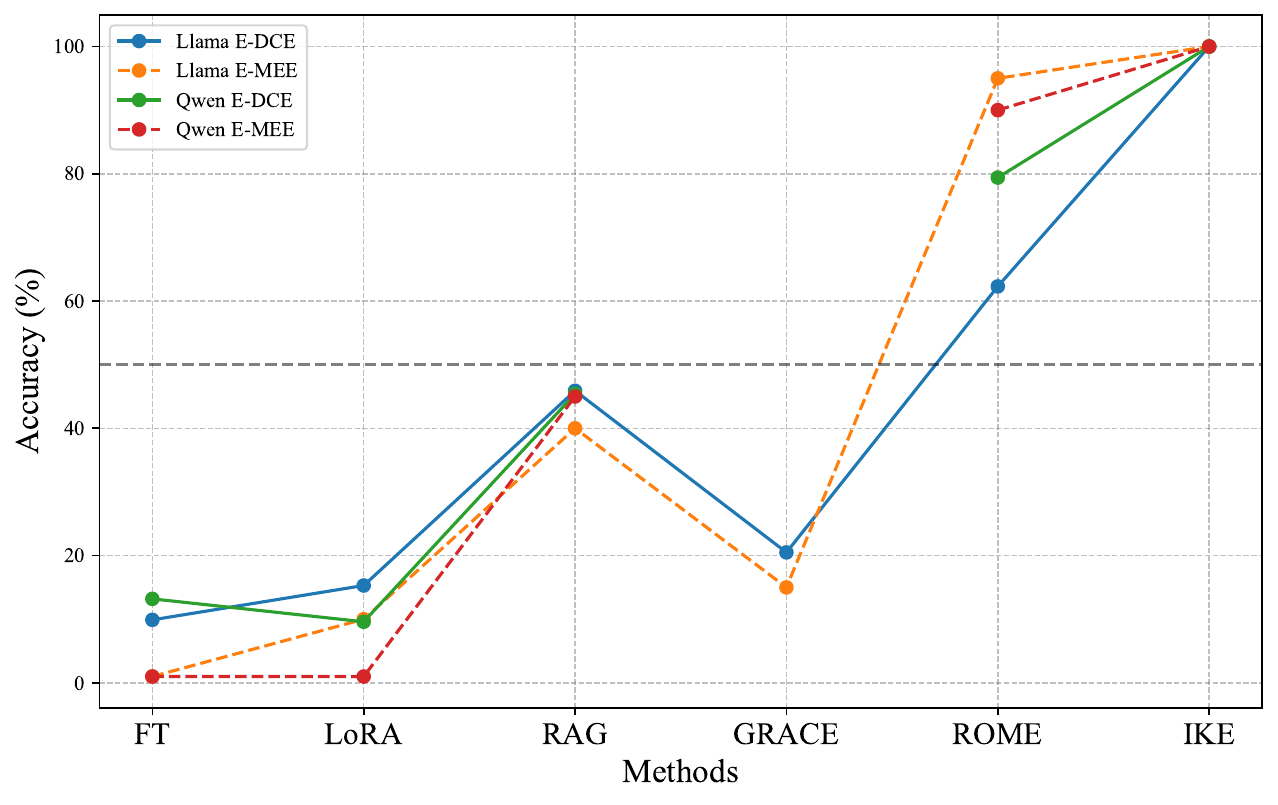}
\caption{A comparison highlighting the question-answering accuracy of six methods—FT,LoRA, RAG, GRACE,ROME and IKE—on the E-DCE and E-MEE of Llama2 7b and Qwen 7b.}
\label{comparative_result}
\end{figure}

Despite the successful deployment of the RE process, comparing our approach with established KE benchmarks such as zsRE~\cite{DBLP:conf/conll/LevySCZ17}, \textbf{E-DCE} and \textbf{E-MEE} introduces significant complexity. This complexity challenges the application of knowledge editing methods and also unveils new features and patterns.
Figure~\ref{comparative_result} illustrates a comparative analysis of six methodologies across the E-DCE and E-MEE, showcasing distinct tiers of performance. 

\par
IKE achieves unparalleled success on both the Llama2 7b and Qwen 7b, with a 100\% effectiveness rate on E-DCE and E-MEE, suggesting its robustness and adaptability to varying complexities. To demonstrate the capabilities of IKE, we selected sample \textit{257} from the E-DCE. This particular sample is challenging because it necessitates accurate identification of four drug combinations and their corresponding effects. According to Appendix~\ref{responses}, both Llama2 7b and Qwen 7b showcased impressive editing capabilities. 
\par
For ROME, Llama2 7b exhibits a relatively robust performance using the ROME  on the E-MEE task, achieving an effectiveness rate of 95\%. However, its performance slightly declines on the more challenging E-DCE task, where it scores 71.5\%. In comparison, Qwen 7b achieves a commendable 79.4\% on the E-DCE  and maintains a solid 90\% effectiveness on the E-MEE.
On the contrary, RAG, LoRA, GRACE, and FT exhibit significantly lower performances, with none surpassing a 50\% effectiveness rate on both datasets.

\subsubsection{Comparison between Structured and Natural Language Knowledge}

Table~\ref{sec:comparative} provides an overview of the comparison between structured and natural language knowledge. While the utilization of structured knowledge for model updates varies among different methods, most methods are less suitable for handling structured knowledge, diverging from their effectiveness in natural language representation. However, with regards to ROME, it is particularly noteworthy that structured knowledge notably improves accuracy. This improvement is associated with a reduction in dimensionality and computational complexity, thereby streamlining the updating process for ROME. We have included comprehensive proofs in Appendix~\ref{proof} to substantiate these advancements.

\section{Conclusion}
In conclusion, our proposed Sequential Fusion method provides a promising solution to the challenges of Few-Shot Domain-Expert Reasoning for updating LLMs. The notable improvements observed in both DCE and our newly developed MEE demonstrate the progress achieved by this approach. During the development of this method, our experiments have shown that guiding LLMs through incremental reasoning for knowledge extraction significantly surpasses the efficacy of few-shot learning techniques. Moreover, our comparative experiments have uncovered new features across different representations and methodologies in knowledge editing within complex texts. This analysis has established the IKE as the superior approach for the second stage of Sequential Fusion. These findings underline the potential of our method to enhance the application of LLMs across various expert domains, contributing to the ongoing development and optimization of domain-specific LLMs.

\section{Limitations}
Despite using GPT-4 with CoT prompting, the model faces challenges in complex numerical reasoning and cognitive tasks, achieving less than 80\% accuracy across two datasets. To improve this, we are exploring a Mixture of Experts (MoE) approach, integrating models like Claude, which have shown improved accuracy in specific cases. However, consistent performance in complex knowledge extraction remains a challenge, indicating the need for further enhancements.


\appendix
\section{Prompts for DCE and MEE}
\label{prompt}
\subsection{Detailed Prompts for DCE}
\subsubsection*{Instruction}
Task Definition is as follows:
\textbf{INPUT}: consists of a sentence, drug mentions within the sentence, and an enclosing context (e.g. paragraph or abstract).
\textbf{OUTPUT}: a set of relations, each consisting of a set of participating drug spans and a relation label
\subsubsection*{Reasoning}
\begin{enumerate}
    \item First, determine the content of the key 'sentence.' If the sentence does not state that the given drugs are used in combination, even if a combination is indicated elsewhere in the wider context, you should output an empty list (\texttt{[]}).
    \item If the sentence indicates that the drugs are used in combination, you should combine it with the content of the key 'paragraph' to determine the effect of the combination. If the effect is positive, you should label it as \texttt{POS}. If the effect is negative, label it as \texttt{NEG}. If the effect is unclear, label it as \texttt{COMB}.
\end{enumerate}
\subsubsection*{Example Outputs}
Here are some output examples,you should output the results in the following format
[{"class": "POS", "spans": [0, 1, 2], "is\_context\_needed": True}]
[{"class": "NEG", "spans": [0, 1], "is\_context\_needed": false}, {"class": "NEG", "spans": [0, 2], "is\_context\_needed": False}]
[{"class": "COMB", "spans": [1, 2, 3], "is\_context\_needed": true}, {"class": "COMB", "spans": [4, 5], "is\_context\_needed": True}]
\subsubsection*{Tips}
Spans are IDs for the combinations of drugs used, and sometimes there may be multiple combinations, such as \texttt{[{"class": "POS", "spans": [0, 2], "is\_context\_needed": true}, {"class": "COMB", "spans": [0, 1], "is\_context\_needed": true}]}". You need to separately assess their effects. The `is\_context\_needed` indicates whether you need to rely on the content of the key 'paragraph' to determine the effects of the drug combinations.

\subsection{Detailed Prompts for MEE}
\subsubsection*{Instruction}
Task Definition is as follows:
\textbf{INPUT}: consists of a sentence, drug mentions within the sentence, and an enclosing context (e.g. paragraph or abstract).
\textbf{OUTPUT}: a set of relations, each consisting of a set of participating drug spans and a relation label

\subsubsection*{Reasoning}
\begin{enumerate}
    \item First, you need to determine the content of the key 'sentence.' If the sentence does not state that the given drugs are used in combination, even if a combination is indicated somewhere else in the wider context, you should output (\texttt{[]}).
    \item Then, if the sentence indicates that the drugs are used in combination, you should combine it with the content of the key 'paragraph' to determine the effect of the combination. If the effect is positive, you should label it as \texttt{POS}. If the effect is negative, you should label it as \texttt{NEG}. If the effect is not yet clear, you should label it as \texttt{COMB}.
\end{enumerate}

\subsubsection*{Example Outputs}
Here are some output examples,you should output the results in the following format:\\
\texttt{[]}\\
\texttt{[{"class": "POS", "spans": [0, 1]}]}\\
\texttt{[{"class": "NEG", "spans": [0, 1]}]}
\subsubsection*{Tips}
The mention of variables may not always correspond exactly with the variable names that appear in the text; a comprehensive judgment based on the content of the text is required.

\section{An example of DCE and MEE}
\label{DCE and MEE}
\subsection{DCE}
\subsubsection*{Input of DCE}
  "sentence": "In patients with NPSLE, the use of high-dose corticosteroids is recommended in combination with immunosuppressants, such as mycophenolate mofetil and intravenous cyclophosphamide pulse therapy."
  "spans": "[ \{"span\_id": 0, "text": "mycophenolate", "start": 126, "end": 139, "token\_start": 19, "token\_end": 20\}, \{"span\_id": 1, "text": "cyclophosphamide", "start": 164, "end": 180, "token\_start": 23, "token\_end": 24\}, \{"span\_id": 2, "text": "corticosteroids", "start": 46, "end": 61, "token\_start": 9, "token\_end": 10\}, \{"span\_id": 3, "text": "mofetil", "start": 140, "end": 147, "token\_start": 20, "token\_end": 21\} ]"
  "paragraph": "Neuropsychiatric Systemic Lupus Erythematosus]. Central nervous system damage, a major organ manifestation of systemic lupus erythematosus (SLE), causes significant morbidity and mortality. Designating this condition as neuropsychiatric SLE (NPSLE), the American College of Rheumatology defines it as involving the central and peripheral nervous systems and being characterized by various manifestations including stroke, seizures, and psychosis. NPSLE treatment mainly seeks to reduce damage accrual. In patients with NPSLE , the use of high-dose corticosteroids is recommended in combination with immunosuppressants , such as mycophenolate mofetil and intravenous cyclophosphamide pulse therapy . This can be accomplished by controlling the activity of the disease, minimizing the use of corticosteroids, and optimizing the management of comorbidities, including cardiovascular risk factors. An international task force analysis of a treat-to-target strategy for SLE (T2T/SLE) recommended targeting remission, preventing damage, and improving quality of life. Thus, more effective and less toxic treatments, such as those using biologics or kinase inhibitors, are still being developed for the treatment of SLE/NPSLE."
\subsubsection*{Output of DCE}
[{"class": "POS", "spans": [0, 2]}, {"class": "POS", "spans": [1, 2]}, {"class": "POS", "spans": [2, 3]"}]

\subsection{MEE}

\subsubsection*{Input of MEE}
"result":
"A 3 (post type) $\times$ 2 (number of posts) 
analysis of variance (ANOVA) for impressions
was estimated. This analysis revealed a 
significant main effect of post type ($F(2,
884)=41.38, p < .001, \eta_p^2=.083$) and 
number of posts ($M_\text{One}=6.17, SD=1.60,
M_\text{Four}=5.74, SD=2.06; F(1, 884)=13.63,
p < .001, \eta_p^2 =.014$).Planned main 
effect contrasts showed that both material 
($F(1, 884)=78.25, p < .001, \eta_p^2=.080$)
and experiential purchase signals ($F(1, 884)
= 7.20, p = .007, \eta_p^2 = .007$) led to 
less positive impressions than non-purchase 
signals did ($M_\text{Non} = 6.51, SD = 
1.65$). Further, experiential purchase 
signals led to more positive impressions 
than material purchase signals ($M_\text{Exp}
= 6.13, SD = 1.72$ vs. $M_\text{Mat} = 5.25, 
SD = 1.96; F(1, 884) = 38.92, p < .001, 
\eta_p^2 = .041$).These main effects were qualified by a significant interaction ($F(2, 884) = 13.93,
p < .001, \eta_p^2 = .028$). Testing H3 
requires the decomposition of the omnibus 2 
df interaction and the use of a planned 1 df
interaction contrast. Using the interaction 
contrast, we find the difference between
experiential and material purchase signals 
was moderated by the number of posts ($F(1,
884) = 13.67, p < .001, \eta_p^2 = .014$).
"spans": [ \{"span\_id": 0, "text": "Signal Type"\} ,\{"span\_id": 1, "text": "Signal Frequency"\} ,\{"span\_id": 2, "text": "Sender Impression"\} ]
"paragraph":
"Study 1 followed a 3 (post: experiential purchase, material purchase, non-purchase) $\times$ 2 (number of posts: 1 vs.4) design. Respondents were told they would evaluate a person based on their Facebook posts. They saw either one or four Facebook posts attributed to their presumed acquaintance Alex Watson.
Participants were recruited from MTurk. Respondents' eligibility was restricted to those residing in the USA with a 95\% or higher approval rate and who had completed at least 50 hits. Given the research context, in this and all other studies, participants were screened to be Facebook users. Respondents were also asked to complete a captcha verification question in order to complete the study.
To prevent multiple responses, participants on MTurk were allowed only one response per MTurk ID. If participants attempted the captcha or the Facebook screener questions multiple times, or entered the study twice for other reasons, only their first answer, even if incomplete, was considered valid. Subsequent responses were excluded from the analysis."
\subsubsection*{Output of MEE}
[{"class": "COMB", "spans": [0, 1]}

\section{binary encoding Strategy} 
\label{sec:binary encoding}
The binary encoding strategy is as follows:

\begin{table}[ht]
\centering
\small
\begin{tabular}{@{}ccccc@{}}
\hline
Category & POS & NEG & COMB & NO\_COMB \\ \hline
Positive Combination & 1 & 0 & 0 & 0 \\
Any Combination & 1 & 1 & 1 & 0 \\ \hline
\end{tabular}
\caption{Mapping Strategy}
\label{tab:mapping}
\end{table}

\section{Evaluation Metrics}
\label{Evaluation Metrics}
The evaluation process involves comparing a generated set $G$ against a standard $S$. An effect score $E$ is assigned where $E=1$ indicates the standard is met, and $E=0$ otherwise.
The details are as follows:
\begin{itemize}
    \item \textit{Combination Score:} Defined by the equation:
    \[ \text{Combination Score} = \frac{\lvert S \cap G \rvert}{\lvert S \rvert} \]
    
    \item \textit{Single Sample Case:} The score \(D\) for a single sample is calculated with the formula: 
    \[ D = \left(\frac{\lvert S \cap G \rvert}{\lvert S \rvert}\right) \times E \]
    
    \item \textit{Multiple Sample Case:} For \(n\) samples, the total score \(T\) is given by:
    \[ T = \sum_{i=1}^{n} \left(\frac{\lvert S_i \cap G_i \rvert}{\lvert S_i \rvert}\right) \times E_i \times \frac{1}{n} \]
    
    \item \textit{Final Accuracy:} Calculated as the ratio of the Total Generative Score to the Total Score.
\end{itemize}
\textbf{Additional Conditions:}
\begin{itemize}
     \item Score is 0 if \(\lvert S \cap G \rvert <= 1\)
    \item Score \(E\) is set to 0 for duplicate effects.
    \item Use the largest \(G\) intersection with \(S\).
    \item Focus on the model's core response accuracy.
\end{itemize}

\section{Comparison of Error Types Across Different Methods}
\label{Error Types}
Table~\ref{types of errors} provides a brief comparison of the error types associated with different methods, highlighting the limitations of LoRA and RAG, and the effectiveness of Sequential Fusion in addressing these challenges.

\begin{table*}[ht]
\centering
\caption{Proportional Distribution of Error Types in Question-Answering Tasks Across Llama2 7b and Qwen 7b on DCE and MEE, calculated as a Percentage of Total Samples.}
\label{types of errors}
\begin{tabular}{cccccccc}
\toprule
\multirow{2}{*}{\centering\textbf{Model}} & \multirow{2}{*}{\centering\textbf{Dataset}} & \multirow{2}{*}{\centering\textbf{Method}} & \multirow{2}{*}{\centering\textbf{No Answers}} & \multirow{2}{*}{\centering\textbf{Contradiction}} & \multirow{2}{*}{\centering\textbf{Error}} & \multirow{2}{*}{\centering\textbf{Omissions}} \\ 
\\
\midrule
\multirow{6}{*}{Llama2 7b} & \multirow{3}{*}{DCE} & LoRA & 17.5 & 32.7 & 30.9 & 1.7 \\
 &  & RAG & 15.1 & 6.3 & 36.2 & 4.8 \\
 &  & Sequential Fusion & 0.0 & 0.0 & 26.8 & 1.5 \\
\cmidrule{2-7}
 & \multirow{3}{*}{MEE} & LoRA & 70.0 & 5.0 & 15.0 & 0.0 \\
 &  & RAG & 10.0 & 5.0 & 60.0 & 0.0 \\
 &  & Sequential Fusion & 0.0 & 0.0 & 15.0 & 0.0 \\
\midrule
\multirow{6}{*}{Qwen 7b} & \multirow{3}{*}{DCE} & LoRA & 64.7 & 2.6 & 16.5 & 0.0 \\
 &  & RAG & 20.2 & 0.0 & 45.6 & 3.6 \\
 &  & Sequential Fusion & 0.0 & 0.0 & 26.8 & 1.5 \\
\cmidrule{2-7}
 & \multirow{3}{*}{MEE} & LoRA & 100.0 & 0.0 & 0.0 & 0.0 \\
 &  & RAG & 10.0 & 0.0 & 55.0 & 0.0 \\
 &  & Sequential Fusion & 0.0 & 0.0 & 15.0 & 0.0 \\
\bottomrule
\end{tabular}
\end{table*}

\section{Exemplary Responses from Llama2 7b and Qwen 7b Models Using IKE}
\label{responses}
\subsection*{Standard answer of E-DCE}
\begin{itemize}
    \item "comb\_drugs":['trastuzumab', 'chemotherapy'],"comb\_effect": "positive"
    \item "comb\_drugs":['pertuzumab', 'chemotherapy'],"comb\_effect": "positive"
    \item "comb\_drugs":['bevacizumab', 'chemotherapy'],"comb\_effect": "positive"
    \item "comb\_drugs":['lapatinib', 'chemotherapy'],"comb\_effect": "positive"
\end{itemize}

\subsection*{Response of Llama2 7b}
A:Trastuzumab and chemotherapy are used in combination, and the effects of the combination are positive. Pertuzumab and chemotherapy are used in combination, and the effects of the combination are positive. Bevacizumab and chemotherapy are used in combination, and the effects of the combination are positive. Lapatinib and chemotherapy are used in combination, and the effects of the combination are positive.Answer: The combination of Trastuzumab and chemotherapy has a positive effect.

\subsection*{Response of Qwen 7b}

A:  Trastuzumab and chemotherapy are combined. The impact of this combination is positive. Pertuzumab and chemotherapy are combined. The impact of this combination is positive. Bevacizumab and chemotherapy are combined. The impact of this combination is positive. Lapatinib and chemotherapy are combined. The impact of this combination is positive .Trastuzumab and chemotherapy have a positive effect when they are combined together.Pertuzumab and chemotherapy have a positive effect when they

\section{ Detailed Proof of Enhanced Accuracy in ROME through Structured Knowledge}
\label{proof}
After structuring the natural language knowledge $\mathcal{P} = {p_1, p_2, ..., p_n}$ into $\mathcal{K} = {k_1, k_2, ..., k_m}$, the objective function for optimizing the value vector $v_*$ in the ROME method can be simplified as:

\begin{small} 
\begin{equation}
\begin{split}
\mathcal{L}(z) = \frac{1}{N}\sum_{j=1}^N \Bigg[&-\log\Pr(o^*|f(x_j+p_j);z) \\
&+ \sum_{i=1}^m D_{\mathrm{KL}}\big(\Pr(k_i|f(p'_j))\big|\Pr(k_i|f(p'_j))_G\bigg]
\end{split}
\end{equation}
\end{small}

Compared to the original natural language form, the structured knowledge form offers the following advantages:

Input dimension reduction
The original function needs to compute probabilities and KL divergences for each $x_j + p$, resulting in a high input dimension
After structuring, only a finite set of $(s, r)$ and $s$ need to be computed, reducing the input dimension to a constant level
Reduced computational complexity
The original function needs to perform autoregressive computation for each $x_j + p$, with a time complexity of $O(N|p|)$
After structuring, only simple lookup is required, reducing the complexity to $O(1)$
Fewer latent variables
In the original function, $x$, $p$, and $p'$ are all latent variables, totaling $N+2$ variables
After structuring, there is only one latent variable $r'$
Simplified probability calculation
The original function needs to model and compute conditional probabilities for the long sequence $x_j+p$
After structuring, only the classification probabilities of $o^*$ and $r'$ need to be computed
The reduction in input dimension and the number of latent variables is evident, as the dimension of structured knowledge $\mathcal{K}$ satisfies:

\begin{align}
\dim(\mathcal{K}) = m \leq n = \dim(\mathcal{P})
\end{align}

and typically $m \ll n$, meaning the dimension of structured knowledge is much smaller than that of natural language knowledge.

The reduction in computational complexity can be explained by the sequence length analysis. Let $L$ be the maximum sequence length, then the complexities of computing probability distributions are:

\begin{align}
\mathcal{O}(\Pr(k_i|f(p'_j))) &\leq \mathcal{O}(L_k) \
\mathcal{O}(\Pr(x|p'j)) &= \mathcal{O}(L{p'})
\end{align}

where $L_k \ll L_{p'}$, since the structured knowledge $k_i$ is usually much shorter than the natural language input sequence $p'_j$.

Finally, the complexities of computing the conditional probability $\Pr(o^*|f(x_j+p_j);z)$ and the classification probability $\Pr(k_i|f(p'_j))$ would also differ.

Through the detailed derivations and analyses provided, we can see that structuring the natural language knowledge can effectively reduce the input dimension, computational complexity, number of latent variables, and the difficulty of probability calculation in the objective function, thereby simplifying the optimization process. This advantage provides a basis for the ROME method to perform semantic editing on model weights more efficiently.

\end{document}